\documentclass[10pt, conference, compsocconf]{IEEEtran}
\usepackage{url}
\usepackage{amsmath}
\usepackage{amssymb}
\usepackage[tight,footnotesize]{subfigure}
\usepackage{algorithm}
\usepackage{algorithmic}
\usepackage{multirow}
\ifCLASSINFOpdf
   \usepackage[pdftex]{graphicx}
\else
\fi
\begin{document}
%
\title{NetDP: An Industrial-Scale Distributed Network Representation Framework for Default Prediction in Ant Credit Pay}


\author{\IEEEauthorblockN{Jianbin Lin, Zhiqiang Zhang, Jun Zhou, Xiaolong Li, Jingli Fang, Yanming Fang, Quan Yu, Yuan Qi}
\IEEEauthorblockA{Ant Financial Services Group\\
Hangzhou, China\\
\{jianbin.ljb, lingyao.zzq, jun.zhoujun, xl.li, sophone.fjl, yanming.fym, jingmin.yq, yuan.qi\}@antfin.com}
}


%


\maketitle

\begin{abstract}
Ant Credit Pay is a consumer credit service in Ant Financial Service Group. Similar to credit card, loan default is one of the major risks of this credit product. Hence, effective algorithm for default prediction is the key to losses reduction and profits increment for the company. 
However, the challenges facing in our scenario are different from those in conventional credit card service. 
The first one is \emph{scalability}. The huge volume of users and their behaviors in Ant Financial requires the ability to process industrial-scale data and perform model training efficiently. 
The second challenges is the \emph{cold-start} problem. Different from the manual review for credit card application in conventional banks, the credit limit of Ant Credit Pay is automatically offered to users based on the knowledge learned from big data. However, default prediction for new users is suffered from lack of enough credit behaviors. It requires that the proposal should leverage other new data source to alleviate the cold-start problem.  

Considering the above challenges and the special scenario in Ant Financial, we try to incorporate default prediction with network information to alleviate the cold-start problem. In this paper, we propose an industrial-scale distributed network representation framework, termed NetDP, for default prediction in Ant Credit Pay. The proposal explores network information generated by various interaction between users, and blends unsupervised and supervised network representation in a unified framework for default prediction problem. Moreover, we present a parameter-server-based distributed implement of our proposal to handle the scalability challenge. Experimental results demonstrate the effectiveness of our proposal, especially in cold-start problem, as well as the efficiency for industrial-scale dataset.

\end{abstract}


%
\IEEEpeerreviewmaketitle

\section{Introduction}
Ant Credit Pay is a consumer credit service in Ant Financial Service Group. Similar to credit card, users are offered with certain credit lines and able to pay for their online/offline shopping with it. Each month, users have to pay their debts before the due day (usually 10th), otherwise default will happen and this could be adverse to users' future loan application. 
Similar to other credit products, loan default is a major risk of Ant Credit Pay, which means default prediction is a key point in risk management. The predicted default probability is one of the most importance factors for admittance management and credit limit grant. Hence, algorithm which makes effective prediction is the key to losses reduction and profits increment for the company.


Ant Credit Pay is an online credit service. In contras with other offline credit card services in conventional bank, it has different business characteristics, which also means that the challenges facing in Ant Credit Pay are different from those in conventional bank. 

The first challenge we are facing is \emph{scalability}. 
In Ant Credit Pay, we serve hundreds of millions of users, which may be tens, or even hundreds, of times larger than the amount of credit card users in a single bank. 
The industrial-scale amount of users and their behaviors requires an industrial-scale data processing and machine learning platform for feature engineering and model training. 
It also requires well-designed distributed algorithms which are able to learn from big data efficiently. 

On the other hand, conventional solutions for default prediction problem tend to learn explainable model (e.g., linear or tree-based model) based on subtle feature engineering. Their performance mainly depends on the effectiveness of input features.
Since most features come from user-provided information and users' behaviors in relative scenarios, the quality of these data decides the effectiveness of feature engineering and the performance of default prediction model. 
Recently, researchers also try to apply new methods, such as deep learning, in default prediction problem. Deep models are able to capture subtle interactions between input features, thus yield better performance. It should be noted that since deep model is still learned from the same feature space as conventional models, its performance also depends on the quality of raw data. 


\begin{figure}[htbp]
\centering
\subfigure[The lifting percentages of default rates of users with certain number of default neighbors (Comparing with users who have 0 default neighbor).]{
  \includegraphics[width=4.0cm]{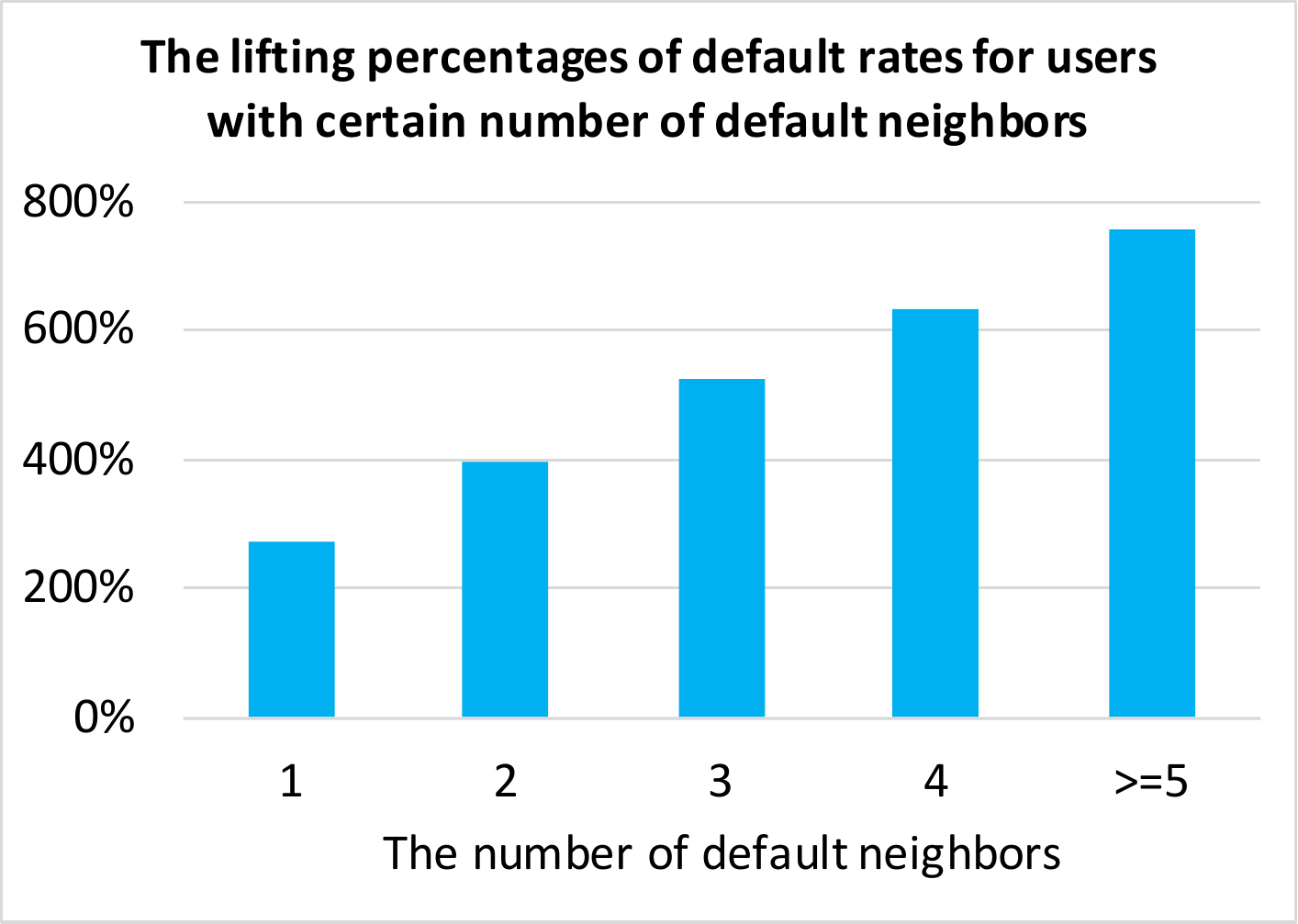}
  \label{fig:lift_of_default_rate}}
\subfigure[The number of neighbors of user groups with different active level.]{
  \includegraphics[width=4.0cm]{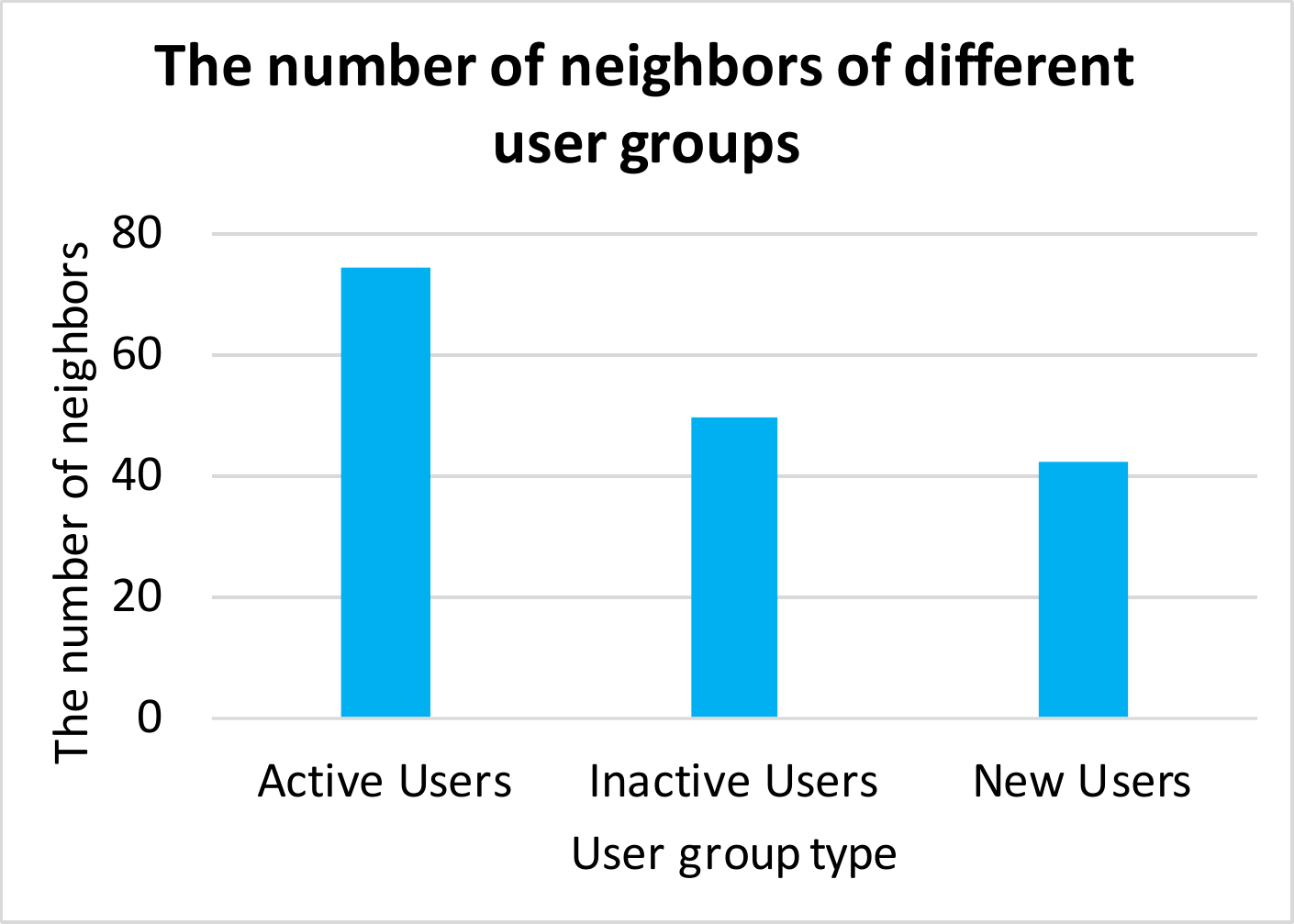}\label{fig:different_user_group}}
\caption{Data Analysis}
\end{figure}

Conventional banks usually manage the credit card applications offline. Their employees manually review applicants' information, which has great contribute to the quality of data. However, in our online service scenario, a huge amount of users stop us from manually reviewing for each application. Our decisions are made according to users' behaviors in relative business in Alipay App (e.g., payment, credit history, etc.). 
Moreover, for those who are inactive in Alipay App, it's hard to acquire high quality data for conventional feature engineering and may result in bad prediction.

Hence here comes the second challenge, the \emph{cold-start} problem. Similar to recommender system, in our scenario, the cold-start problem means the challenge in predicting default probability for inactive users or new users, due to the lack of enough data. Hence, new data source as well as new algorithm need to be applied to alleviate the cold-start problem. 


In Ant Financial, a user will interact with other users in various kinds of businesses (e.g., social relations, fund interactions, common interests, etc.). It is natural to build a social network, in which user acts as node and their interaction acts as edge. In real world, it is easy to find that people with different level of credit risk tend to interact with different crowd. We demonstrate this observation in Figure \ref{fig:lift_of_default_rate}. We first calculate the default rates of user groups with certain number of default neighbors respectively, and then show the lifting percentages of default rates comparing with users who have none default neighbor. As shown in Figure \ref{fig:lift_of_default_rate}, the more default neighbors a user have, the higher default rate he will be. Specially, the credit risk of users with no less than 5 default neighbors is almost $800\%$ higher than who never interact with default user. The \emph{aggregation} of users with similar risk in social network implies that the structural information of social relations can be beneficial to the default prediction problem. 

On the other hand, users who rarely participate in credit-relative businesses may still have rich social interactions. Figure \ref{fig:different_user_group} demonstrates the average amount of neighbors of user groups with different active level. New users mean those who have signed up in Alipay App in a month, while other users are divided into two groups (i.e., active users and inactive users) based on their frequency of using the App. The active users has the most neighbors due to the higher frequency of use. But for both inactive users and new users, more than 40 neighbors are found in the social network, which are enough to learn high-quality network representations for them. 

However, conventional methods for default prediction rarely utilize network structural information effectively. It's hard to find literatures about applying network structural information to improve the default prediction problem, especially in such an industrial scenario with billions of users and tens of billions connections between users.

Considering the above challenges and the special scenario in Ant Financial, we present \textbf{NetDP} (DP is short for \textbf{D}efault \textbf{P}rediction), an industrial-scale distributed network representation framework for default prediction in Ant Credit Pay. NetDP is a flexible framework which supports both unsupervised and supervised network representation simultaneously. The unsupervised module tends to depict the global structural information in the whole network, while supervised module is responsible for modeling local structural information of labeled data. Then, the ensemble module applies \textbf{M}ultiple \textbf{A}dditive \textbf{R}egression \textbf{T}ree (MART) to blend the output of unsupervised and supervised model, and assigns a final predicted default probability for certain user. 

Thanks to the succinct modeling and efficient distributed implement, NetDP can modeling the structural information of a social network with billions of nodes and tens of billions of edges in several hours. To our best knowledge, there is not a published method which can learn representation in a network of such magnitude efficiently. Moreover, experimental results show that the proposal can actually improve the performance of default prediction, especially for new users.

The rest of this paper is organized as follow: section 2 give a preliminary of the proposal. Some notations used in the following sections are defined here. Section 3 presents NetDP, the proposed distributed network representation framework for default prediction. Unsupervised module, supervised module, as well as the ensemble module are introduced respectively. We also present details about the distributed implement of NetDP. Section 4 shows the experimental settings and results to demonstrate the performance of our proposal in default prediction problem, especially in alleviating the cold-start problem. We also show the efficiency of our distributed implement. Section 5 is the related works about the default prediction problem and our proposal. And we make a conclusion of this work in the last section.

\section{Preliminary}
Given a directed network $G=\{V, E\}$, in which $V=\{v_i\}_{i=1}^{N}$ denotes the node set of size $n$ and $E=\{(v_i, v_j)\}$ denotes the directed edge set (a pair $(v_i, v_j)$ represents the edge from node $v_i$ to node $v_j$).  Let $\mathcal{N}_{v_i}=\{v_j|(v_i, v_j) \in E\}$ denotes the set of neighbor nodes of $v_i$. In our scenario, a node represents a user of Alipay and an edge $(v_i, v_j)$ represents that user $v_i$ interacts with $v_j$. The interactions between users include social friend relationship, fund transfer between users, transaction between buyer and seller, etc. To simplify the problem, different types of interactions are treated in a unified manner. Irrelevant or not strong enough interactions between users are dropped in case the useless or even pernicious noises are brought. 

The unsupervised representation learning module will assigned a learnt $d$-dimensional vector $\mathbf{u}^u_i \in \mathbb{R}^d$ (the superscript $u$ indicates this representation is generated by \textbf{u}nsupervised learning) to each node $v_i$, which represents the global structural information of $v_i$. A little portion of users who have paid with Ant Credit Pay are labeled according to whether they defaulted or not. The label $y_i=1$ means $v_i$ have defaulted on a loan, otherwise $y_i=0$. After model training, the supervised module will assign a learnt $k$-dimensional vector $\mathbf{u}^s_i \in \mathbb{R}^k$ (the superscript $s$ indicates this representation is generated by \textbf{s}upervised learning) to represent the local structural information of $v_i$, as well as a  score $\hat{y}_i \in (0,1)$ to $v_i$ as the predicted default probability.

\section{The proposed method: NetDP}

In this section, we first introduce the overall framework of the proposed NetDP briefly, then give detailed formalizations of unsupervised representation learning module and supervised representation learning module. At last, we will present the efficient distributed implementation of each component in NetDP.

\begin{figure}[th]
\includegraphics[scale=0.15]{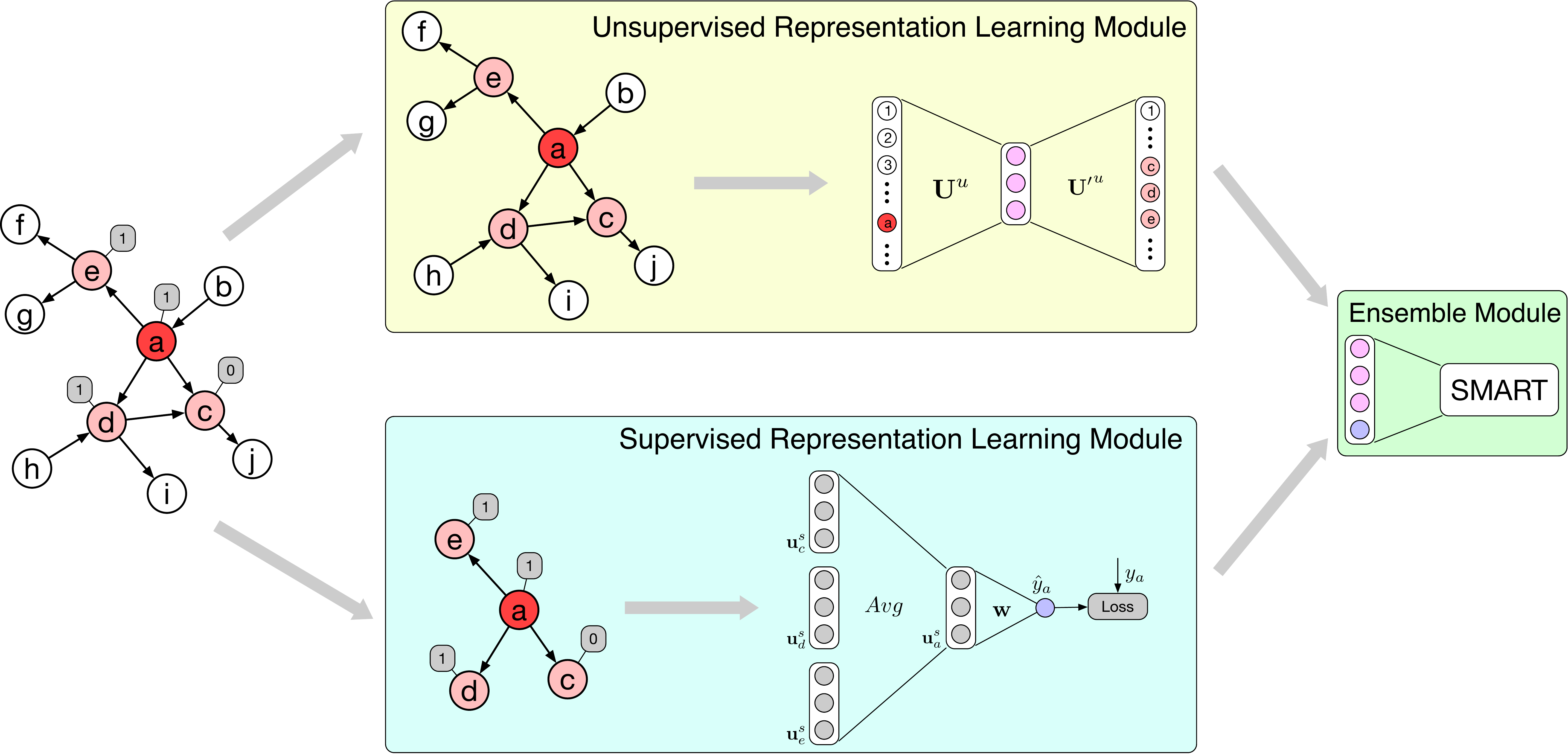}
\caption{The overall framework of NetDP.}
\label{fig:netdp_framework}
\end{figure}

\subsection{Overall Framework}
Figure \ref{fig:netdp_framework} demonstrates the overall framework of NetDP. The input includes two parts: a directed network $G$ consisting of Alipay users and the social interactions between them, and the default labels tagged on a little portion of users (gray blocks tagged on nodes with 0 or 1 inside in Figure \ref{fig:netdp_framework}). The whole network without labels acts as the input of \emph{unsupervised representation learning module} and the learned representations (i.e., $\mathbf{u}^u$) for each node are outputted to ensemble module. The \emph{Supervised representation learning module} takes the labeled nodes and their neighbors as input. It produces learned representations (i.e., $\mathbf{u}^s$) for each node and the predicted default probability (i.e., $\hat{y}$). The unsupervised representation vector $\mathbf{u}^u$ and the supervised predicted score $\hat{y}$ are concatenated as $(d+1)$-dimensional vector, which becomes the input of the \emph{ensemble module}. A distributed \textbf{M}ultiple \textbf{A}dditive \textbf{R}egression \textbf{T}ree (MART) is applied to ensemble the output of unsupervised and supervised module. A final predicted default probability is given to represent the credit risk of certain user.

\subsection{Formalization of Network Representation Learning}
Our goal is to predict whether a user will default or not based on the structural information learned from the given social network $G$. To achieve this, NetDP performs unsupervised and supervised network representation learning simultaneously. 
On the one hand, the unsupervised method, which takes the whole network as input, can learn effective representation to encode the global structural information of each node. Without biased adjustment by supervised information, the unsupervised representations can objectively reflect the structural characteristics of nodes. On the other hand, the default prediction problem is still a supervised learning task. Hence, we design a supervised network representation method to capture the local structural information of labeled nodes by focusing on modeling the relationship between labeled nodes and their neighbors. We will introduce the formalizations of these two methods in the following subsections.

\subsubsection{The unsupervised method}
Many state-of-the-act unsupervised network representation methods represent a node as a vector of the low-rank hidden space, which encodes the structural information of the corresponding node among the whole network. A common assumption of these works is that closely connected nodes should be close to each other in the low-rank hidden space. 
In our case, closely connected users tend to have similar credit risk (Figure \ref{fig:lift_of_default_rate}). Hence, the ability of encoding closely connected nodes into close low-rank vectorized representations is the key point to build default prediction model with unsupervised network representation.


Some recent works (\cite{perozzi2014deepwalk, grover2016node2vec, tang2015line}) provide a new direction to capture structural information from social network. These methods learn low-rank vectors to represent each node, which are able to preserve relations between nodes in the network. DeepWalk \cite{perozzi2014deepwalk} and Node2vec \cite{grover2016node2vec} have two main steps: apply random walk to generate the node sequences, and then perform skip-gram model to generate representations for each node.

Random walk is utilized to extract the high-order topological information from the network. At each walk, a neighbor of the last visited node is samples uniformly until the maximum walking step is reached. 
Random walk is efficient only if it's able to fit the whole network into the memory of a single machine. In the single machine case, the lookup of neighbors for a certain node can perform much faster because no machine communication would happen during the lookup procedure. However, the situation becomes challenging when the network is too large that it has to be partitioned and stored on several machines. If a node and its neighbors are not stored in the same machine, the time of neighbors lookup procedure will increase from $O(l)$ to $O(2l + 2c)$, where $l$ denotes the time of lookup in the same machine, and $c$ denotes the communication time between two machines. $2l$ means one local lookup and one remote lookup are required, while $2c$ means one communication to issue the lookup command from the local machine to the remote one and one  communication to aggregate the lookup results. In general, $c$ is quite larger than $l$. Therefore, random walk become inefficient if the whole network is too large to store in the memory of a single machine. 

In our scenario, the social network contains billions of users and tens of billions of interactions between them. It has to be stored in several machines. It's impossible to perform the random-walk based or higher order proximity based network representation methods in such a huge network efficiently. On the other hand, the social network in Alipay App has rich enough first order proximity (i.e., more than 40 neighbors of each node) for modeling network representation. Therefore, in our proposed unsupervised method, only the direct neighbors are sampled to optimize the representation of the target node. 

In our algorithm, a node $v_i$ is represented as a $d$-dimensional vector $\mathbf{u}^u_i$. At each time, for a target node $v_i$, several neighbors of $u_i$ are sampled from its neighbor set (i.e., $\{v_j\} \sim \mathcal{N}_{v_i}$). We denote the log-likelihood of co-occurrence of $v_i$ and $v_j$ as:
\begin{equation} \label{eq:un_likelihood}
\log{P(v_j|v_i)} = \log\frac{\exp({\mathbf{u}^u_i}^{\top} \mathbf{u}^u_j)}{\sum_{v_k \in V}{\exp({\mathbf{u}^u_i}^{\top} \mathbf{u}^u_k)}}.
\end{equation}
To preserve the structural information of the social network, the representations of nodes are optimized to minimize the negative log-likelihood of co-occurrences as follow:
\begin{equation} \label{eq:un_loss}
\mathcal{L}^u = -\sum_{v_i \in V}{\sum_{v_j \in \mathcal{N}_{v_i}}{\log{P(v_j|v_i)}}}
\end{equation}

In order to optimize the loss efficiently, negative sampling  technique is applied and the loss can be revised as:
\begin{equation} \label{eq:neg_samp}
\mathcal{L}^u \approx -\sum_{v_i \in V}{\sum_{v_j \in \mathcal{N}_{v_i}}{\log(\sigma({\mathbf{u}^u_i}^\top \mathbf{u}^u_j)) + \sum_{v_k \sim V}{\log(\sigma({\mathbf{u}^u_i}^\top \mathbf{u}^u_k))} }}, 
\end{equation}
where $\sigma(\cdot)$ is the sigmoid function. 

By minimizing the above loss, the learned node representations can preserve the structure information of the whole network, which is beneficial to predict users' credit risk. 

\subsubsection{The Supervised Method}
Although the network representations generated by unsupervised method can be treated as input feature of a classifier to recognize whether a user will default or not, an end-to-end supervised network representation method is still needed in this problem. It's easy to observe that the credit risk of a certain user can be reflected by the credit risk of his/her neighbors (Figure \ref{fig:lift_of_default_rate}), which means the local labeled information is beneficial to default prediction. 

Based on this observation, the proposed supervised method represents a target user with the aggregation of his/her neighbors' representations. That is, at $k$-th step, the representation of a target node is a non-linear transformation of the average of his/her neighbors' representation in $k-1$ step:
\begin{equation} \label{eq:su_node_rep}
{\mathbf{u}^s_i}^{(k)} = \sigma(\mathbf{W}_1 \sum_{v_j \in \mathcal{N}_{v_i}}{\frac{{\mathbf{u}^s_j}^{(k-1)}}{|\mathcal{N}_{v_i}|}}),
\end{equation}
where $\mathbf{W}_1 \in \mathbb{R}^{k \times k}$ is a trainable matrix. The prediction of default probability is given based on the node representation. 
\begin{equation} \label{eq:su_prediction}
\hat{y}_i = \sigma({\mathbf{w}_2}^{\top} \mathbf{u}^s_i), 
\end{equation}
where $\mathbf{w}_2 \in \mathbb{R}^k$ is another trainable vector. At last, $\mathbf{u}^s$, $\mathbf{W}_1$ and $\mathbf{w}_2$ are optimized to minimize the cross entropy loss between the prediction $\hat{y}$ and the ground truth $y$ of each labeled data.
\begin{equation} \label{eq:su_prediction}
\begin{aligned}
\mathcal{L}_s & = - \sum_{v_i \in V}{y_i \log{\hat{y}_i} + (1-y_i)\log{(1-\hat{y}_i)}} \\
& \qquad + Reg(\mathbf{u}^s, \mathbf{W}_1, \mathbf{w}_2),
\end{aligned}
\end{equation}
where $Reg(\mathbf{u}^s, \mathbf{W}_1, \mathbf{w}_2)$ is the $l_2$-norm regularization of all trainable parameters. 
Different from the representations generated by unsupervised method, the supervised representations pay more attention to the local structural information of the labeled data. The predicted score $\hat{y}$ is utilized in the ensemble module as one of the input features for MART training.

\subsection{Distributed Implementations}
In order to learn node representations in such an industry-scale social network efficiently, we implement and deploy the proposed NetDP on \emph{KunPeng} platform \cite{zhou2017kunpeng}. KunPeng is a high-performance distributed machine learning platform, which provide parameter-server-based API for implementing parallel machine learning algorithms to learn from industrial-scale data. 

In the proposed NetDP, to optimize the representation of a target node, we only need to lookup the representations of its direct neighbors. In this situation, the social network $G$ has to be formed as adjacency lists (i.e., each record stores one target node and all of its neighbors together) first. Then the whole set of adjacency lists is divided into serval parts and stored in the memory of serval machines. Hence, the neighbor lookup procedure for a certain node will only happen in one machine, which contributes to shorten the communicational time between different machines and make the training procedure efficient. 

Moreover, we implement the parallel mini-batch \textbf{S}tochastic \textbf{G}radient \textbf{D}escent (SGD) to accelerate the training procedure. Mini-batch SGD solves optimization problem iteratively. At each step, each worker randomly selects a mini-batch of nodes, and retrieves the neighbors of them. It then computes the gradients of the objective function with respect to different trainable parameters, and do parameter update. The training procedure goes on until convergence happens or the maximum epoch is reached. Algorithm \ref{alg:un} demonstrates the distributed implementation of unsupervised network representation in NetDP. The implementation of supervised method is similar to Algorithm \ref{alg:un}. 

\begin{algorithm}[ht]
    \caption{Distributed implementation of Unsupervised Network Representation}
    \label{alg:un}
    \begin{algorithmic}[1]
    \REQUIRE ~~\\
        $G$: the social network\\
        $m$: the maximum epoch to stop optimization
    \ENSURE ~~\\
        $\mathbf{U}^u$: matrix with columns as node representations\\
        
    \STATE Transform $G$ to adjacency list format and partition it into several servers randomly.
    \STATE Initialize $\mathbf{U}^u$ randomly. 
    \STATE $i = 0$
    \WHILE {$i < m$}
        \STATE \textbf{in worker $k$}:
        \STATE \qquad Shuffle node set in this worker $V_k^{(i)}$.
    	\STATE \qquad \textbf{for} {mini-batch $b$ in $V_k^{(i)}$} 
		\STATE \qquad \qquad Pull the adjacency lists of $b$ from servers.
		\STATE \qquad \qquad Pull relative representations from servers.
		\STATE \qquad \qquad Calculate gradients of $\mathcal{L}_u$.
		 \STATE \qquad \qquad Update local representations. 
		 \STATE \qquad \qquad Push the local representations to servers.
	\STATE \qquad \textbf{end for}
	\STATE Synchronization.
	\STATE $i = i + 1$
    \ENDWHILE
    \end{algorithmic}
\end{algorithm}

%

\begin{table}[t]
\label{tbl:network}
\centering
  \begin{tabular}{clclclc}
  \hline
  \# of Users  & \# of relations  & Avg. of in degree & Avg. of out degree \\
  \hline
  2.54b & 58.8b & 57.2 & 48.3 \\
  \hline
  \end{tabular}
  \caption{Statistics of the social network}
\end{table}

\begin{table}[t]
\label{tbl:labeled_data}
\centering
  \begin{tabular}{clclclc}
  \hline
  \# of Users  & \# of Default Users  & The First Month & The Last Month \\
  \hline
  20m & 2.04m & 201703 & 201711 \\
  \hline
  \end{tabular}
  \caption{Statistics of the labeled data}
\end{table}

\section{EXPERIMENTS}
In this section, we demonstrate the effectiveness of NetDP in solving default prediction problem, especially the cold-start problem, as well as its efficiency in handling industrial dataset.

\subsection{Experimental Settings}
The dataset consists of two parts: the social network and the labeled data. Table I shows the statistic information of the social network. Around 1.04 billions of users and 58.5 billions of relations between them are involved in the social network. To our best knowledge, this could be one of the biggest social networks mentioned in recent literatures, and none of the state-of-the-art network representation methods have reported results on a social network with similar scale. 

Table II shows the statistic information of the labeled data. We drop lots of the users who never default before to keep the default rate at around $10\%$. The labeled data are collected from March, 2017 to November, 2017. We divide them into two parts: the training set from March, 2017 to July, 2017, and the test set from August, 2017 to November, 2017. We train NetDP with the training set and report the performance of test set as results. 

Three results are reported as follow:
\begin{itemize}
	\item \textbf{NetDP}: the output of the ensemble module (MART) of NetDP.
	\item \textbf{BenchDP}: the output of a conventional default prediction model based on credit-relative features. 
	\item \textbf{NetDP+BenchDP}: a weighted average of the output of NetDP and BenchDP. The weights of the two models are adjusted based on training set. 
\end{itemize}

We utilize the \textbf{K}olmogorov-\textbf{S}mirnov statistic (KS) as metric to demonstrate the performance of default prediction. The higher value of KS statistic means the better performance. 


\begin{figure*}[htbp]
\centering
\subfigure[KS performance of different methods]{
	\includegraphics[width=4.2cm]{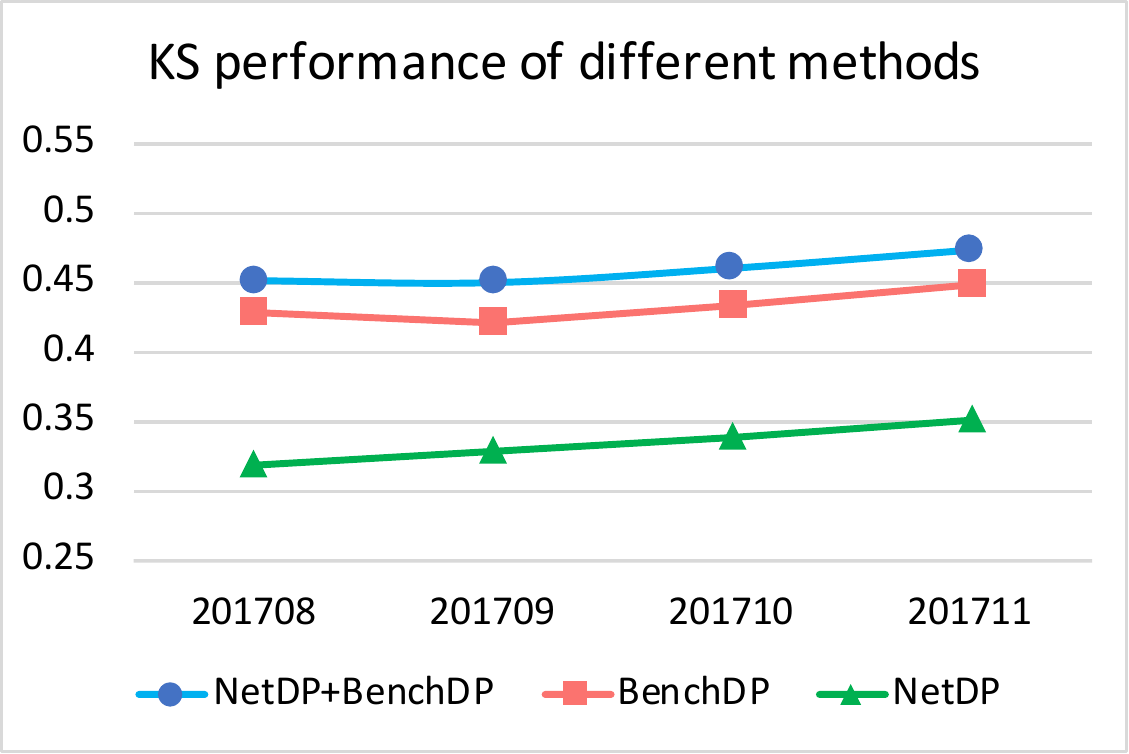}
\label{fig:overall_performance}}
\subfigure[KS performance on active users]{
  \includegraphics[width=4.2cm]{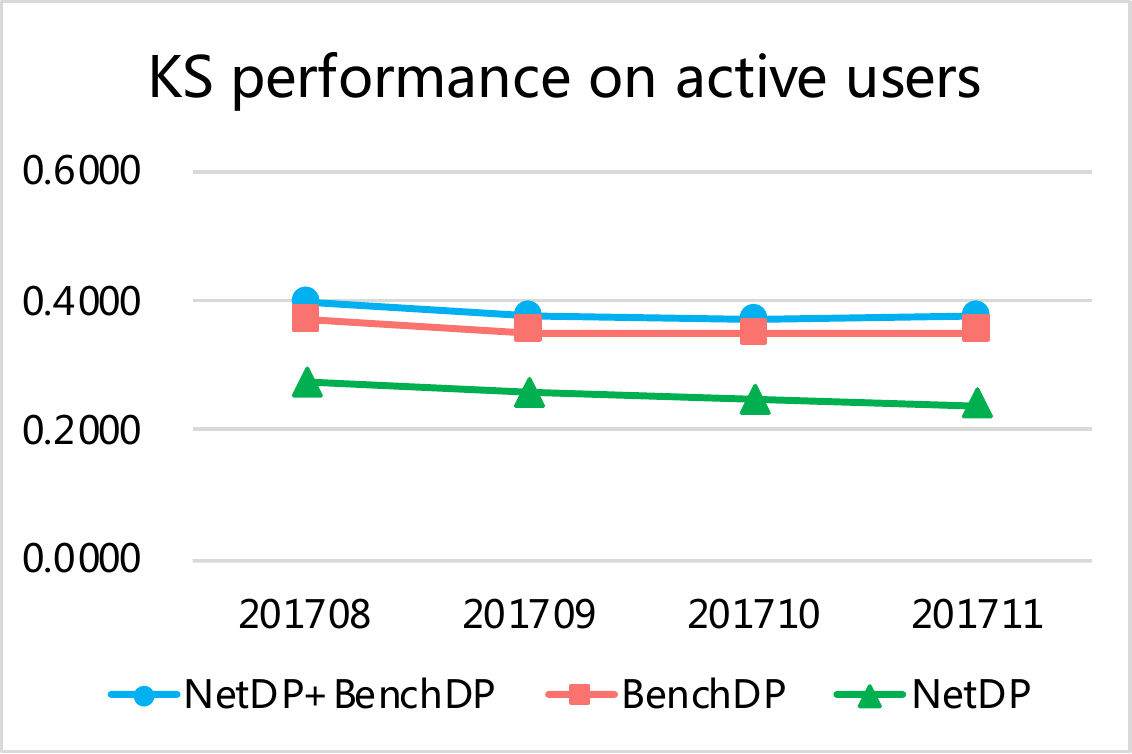}\label{fig:active_performance}}
\subfigure[KS performance on inactive users]{
  \includegraphics[width=4.2cm]{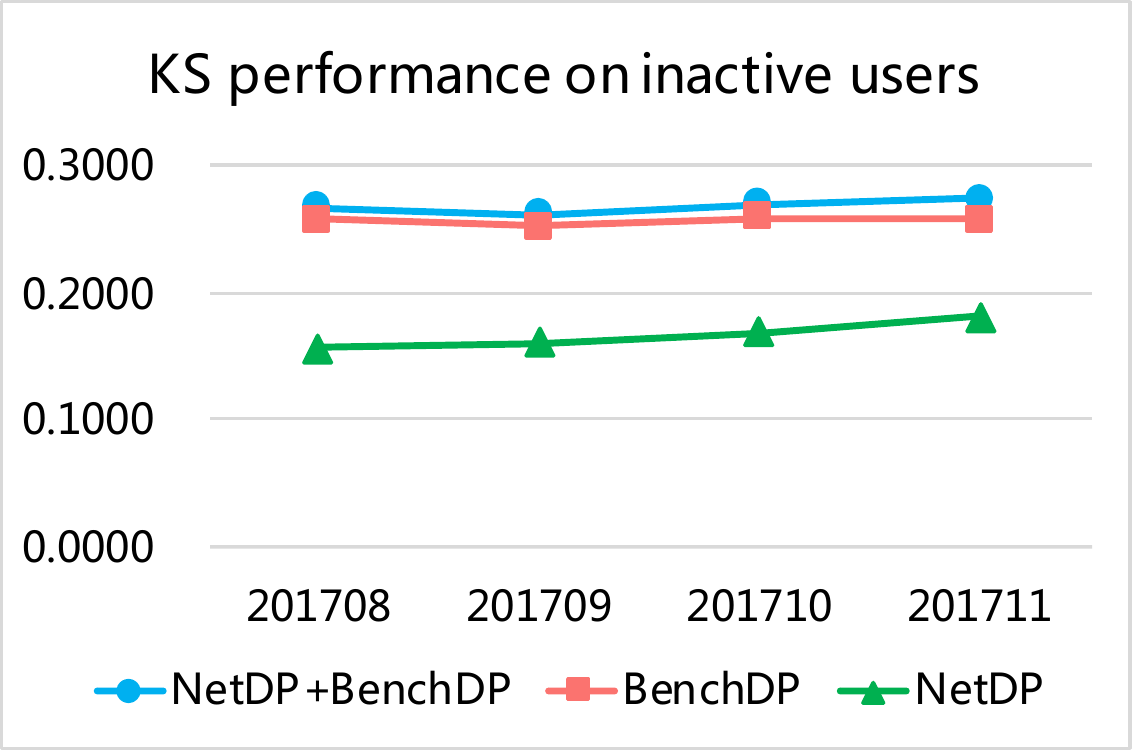}\label{fig:inactive_performance}}
\subfigure[KS performance on new users]{
  \includegraphics[width=4.2cm]{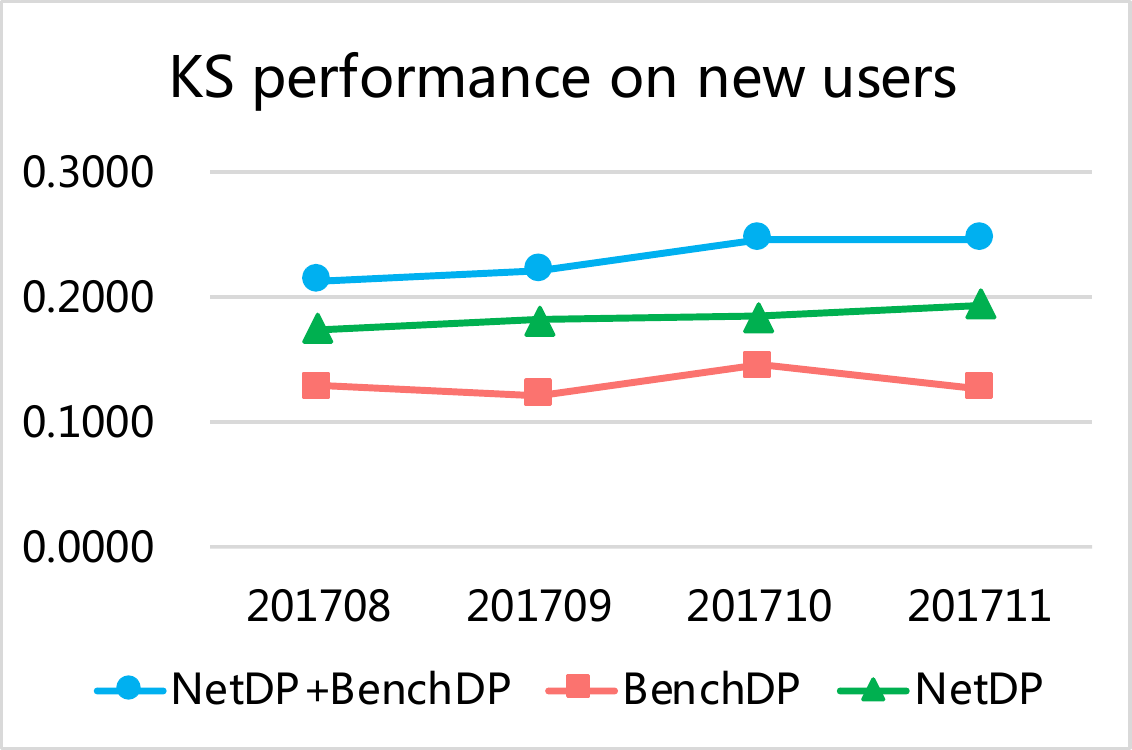}\label{fig:new_performance}}
 \label{fig:cold_start}
\caption{KS performance on different type of user groups}
\end{figure*}

\subsection{Experimental Results}
As shows in Figure \ref{fig:overall_performance}, BenchDP outperforms NetDP. That's because the conventional model is based on well-designed credit features, while NetDP only models the network structural information without any features. However, a simple weighted average result (i.e., NetDP+BenchDP) can achieve significant improvement in comparing with BenchDP, which means the structural information is beneficial to the conventional models. It also illustrates that the proposed NetDP has the ability to capture effective structural information in an  industrial-scale social network.

Moreover, to demonstrate the ability of NetDP in alleviating cold-start problem, we evaluate KS performance of different methods in different types of user groups (Figure 6). In Figure \ref{fig:active_performance} and \ref{fig:inactive_performance}, BenchDP still outperforms NetDP thanks to the rich enough credit features, but the distance diminishes in inactive users comparing to active users. However, the situation become different in Figure \ref{fig:new_performance}. For new users, NetDP outperforms BenchDP and the weighted average of NetDP and BenchDP achieve great improvement than it on active or inactive users. These experimental results prove the effectiveness of utilizing network representation methods to predict for those who have less credit information. 

We have implemented and deployed NetDP in the data center of AlibabaCloud. The unsupervised network representation module has to apply around 1000 cpu cores for modeling network mentioned above, and finishes training in around 5 hours 40 minutes. The supervised module acquires 20 cpu cores and perform training in 1 hour. In summary, the proposed NetDP is quite efficient in modeling industrial-scale social networks.

\section{Related work}
Conventional statistical techniques including linear discriminant analysis, logistic regression and naive bayes method have been widely used for building loan default prediction model \cite{thomas2017credit}. More advance models, such as \textbf{S}upport \textbf{V}ector \textbf{M}achine (SVM) and neural networks are introduced as promising data mining tools,  which provide an alternative to statistical techniques in building default prediction models \cite{lessmann2015benchmarking,abdou2016predicting,louzada2016classification}.

%

Recently, the network representation model plays an increasingly important role to encode an existing network into a low-rank representation space to facilitate network structure analysis 
Our work is mainly related to structure based methods. \cite{perozzi2014deepwalk} first deployed truncated random walks on networks to generate node sequences, and then leverage skip-gram model to learn node representations. \cite{grover2016node2vec} designed a biased random walk to balance breadth-first sampling and depth-first sampling. \cite{tang2015line} designed objective functions to preserve the first-order proximity and second-order proximity. Although these methods are able to scale up to large dataset in a single machine, it is still necessary to proposed an efficient distributed network representation method which is able to handle industrial-scale networks.

\section{Conclusion}
This paper aims to improve the prediction of loan default in Ant Credit Pay. In order to overcome the scalability and cold-start challenges, we propose to incorporate default prediction with social network information, and present NetDP, an industrial-scale distributed network representation framework, to learn the structural information of users in a very-large social network. The proposal models global structural information with unsupervised network representation methods, as well as local structural information with supervised network representation method, and blends the outputs with MART for final prediction. We also present a distributed implementation of NetDP based on KunPeng platform, which it's able to learn representation from a social network with billions of users and tens of billions of relations. The experimental results shows that with the help of NetDP, conventional default prediction model can achieve better performance, especially in cold-start users.  

\bibliographystyle{IEEEtran}
\bibliography{IEEEabrv,cite}

\begin{thebibliography}{1}
\providecommand{\url}[1]{#1}
\csname url@samestyle\endcsname
\providecommand{\newblock}{\relax}
\providecommand{\bibinfo}[2]{#2}
\providecommand{\BIBentrySTDinterwordspacing}{\spaceskip=0pt\relax}
\providecommand{\BIBentryALTinterwordstretchfactor}{4}
\providecommand{\BIBentryALTinterwordspacing}{\spaceskip=\fontdimen2\font plus
\BIBentryALTinterwordstretchfactor\fontdimen3\font minus
  \fontdimen4\font\relax}
\providecommand{\BIBforeignlanguage}[2]{{%
\expandafter\ifx\csname l@#1\endcsname\relax
\typeout{** WARNING: IEEEtran.bst: No hyphenation pattern has been}%
\typeout{** loaded for the language `#1'. Using the pattern for}%
\typeout{** the default language instead.}%
\else
\language=\csname l@#1\endcsname
\fi
#2}}
\providecommand{\BIBdecl}{\relax}
\BIBdecl

\bibitem{perozzi2014deepwalk}
B.~Perozzi, R.~Al-Rfou, and S.~Skiena, ``Deepwalk: Online learning of social
  representations,'' in \emph{Proceedings of the 20th ACM SIGKDD international
  conference on Knowledge discovery and data mining}.\hskip 1em plus 0.5em
  minus 0.4em\relax ACM, 2014, pp. 701--710.

\bibitem{grover2016node2vec}
A.~Grover and J.~Leskovec, ``node2vec: Scalable feature learning for
  networks,'' in \emph{Proceedings of the 22nd ACM SIGKDD international
  conference on Knowledge discovery and data mining}.\hskip 1em plus 0.5em
  minus 0.4em\relax ACM, 2016, pp. 855--864.

\bibitem{tang2015line}
J.~Tang, M.~Qu, M.~Wang, M.~Zhang, J.~Yan, and Q.~Mei, ``Line: Large-scale
  information network embedding,'' in \emph{Proceedings of the 24th
  International Conference on World Wide Web}.\hskip 1em plus 0.5em minus
  0.4em\relax International World Wide Web Conferences Steering Committee,
  2015, pp. 1067--1077.

\bibitem{zhou2017kunpeng}
J.~Zhou, X.~Li, P.~Zhao, C.~Chen, L.~Li, X.~Yang, Q.~Cui, J.~Yu, X.~Chen,
  Y.~Ding \emph{et~al.}, ``Kunpeng: Parameter server based distributed learning
  systems and its applications in alibaba and ant financial,'' in
  \emph{Proceedings of the 23rd ACM SIGKDD International Conference on
  Knowledge Discovery and Data Mining}.\hskip 1em plus 0.5em minus 0.4em\relax
  ACM, 2017, pp. 1693--1702.

\bibitem{thomas2017credit}
L.~Thomas, J.~Crook, and D.~Edelman, \emph{Credit scoring and its
  applications}.\hskip 1em plus 0.5em minus 0.4em\relax Siam, 2017, vol.~2.

\bibitem{lessmann2015benchmarking}
S.~Lessmann, B.~Baesens, H.-V. Seow, and L.~C. Thomas, ``Benchmarking
  state-of-the-art classification algorithms for credit scoring: An update of
  research,'' \emph{European Journal of Operational Research}, vol. 247, no.~1,
  pp. 124--136, 2015.

\bibitem{abdou2016predicting}
H.~A. Abdou, M.~D.~D. Tsafack, C.~G. Ntim, and R.~D. Baker, ``Predicting
  creditworthiness in retail banking with limited scoring data,''
  \emph{Knowledge-Based Systems}, vol. 103, pp. 89--103, 2016.

\bibitem{louzada2016classification}
F.~Louzada, A.~Ara, and G.~B. Fernandes, ``Classification methods applied to
  credit scoring: Systematic review and overall comparison,'' \emph{Surveys in
  Operations Research and Management Science}, vol.~21, no.~2, pp. 117--134,
  2016.

\end{thebibliography}

\end{document}